# Epidemic Modeling with Generative Agents

Ross Williams[1], Niyousha Hosseinichimeh[1], Aritra Majumdar[2], Navid Ghaffarzadegan[1]*

[1]Industrial and Systems Engineering, Virginia Tech; Falls Church, VA 22043, USA.
[2]Computer Science, Virginia Tech; Falls Church, VA 22043, USA.

*Corresponding author. Email: navidg@vt.edu

**Abstract**

This study offers a new paradigm of individual-level modeling to address the grand challenge of incorporating human behavior in epidemic models. Using generative artificial intelligence in an agent-based epidemic model, each agent is empowered to make its own reasonings and decisions via connecting to a large language model such as ChatGPT. Through various simulation experiments, we present compelling evidence that generative agents mimic real-world behaviors such as quarantining when sick and self-isolation when cases rise. Collectively, the agents demonstrate patterns akin to multiple waves observed in recent pandemics followed by an endemic period. Moreover, the agents successfully flatten the epidemic curve. This study creates potential to improve dynamic system modeling by offering a way to represent human brain, reasoning, and decision making.

**One-Sentence Summary:** A new modeling technique using generative AI applied to an epidemic to incorporate human reasoning and decision making.

# I. Introduction

The COVID-19 pandemic has emphasized the vital role of human behavior, including practices such as social distancing, wearing masks, and vaccination, in shaping the trajectory of the disease (1, 2). People's responses to perceived risks create a feedback loop that influences the pandemic's future course, which in turn influences pandemic outcomes and future public risk perception (3). Despite widespread recognition of the importance of incorporating changes in human behavior in epidemic models (4-6), many modelers have yet to comprehensively address this crucial aspect (6, 7). While some empirically grounded models have integrated past human behavior data to model disease progression, often by exogenously feeding time series data (e.g., mobility and seasonality data) (8), only a limited number of models have endogenously formulated the dynamic nature of human behavior as a response to the state of the disease (9, 10). The lack of endogenous formulations have resulted in biases in long-term projections (10) and hence to suboptimal policy recommendations (11). Advancing the inclusion of human behavior in epidemic models is, therefore, essential to improve the accuracy of projections and inform more effective policy decisions (3).

Incorporating human behavior into epidemic models entails multiple challenges (12). Human behavior is complex and diverse, varying across individuals, communities, cultures, and contexts. Integrating such complexity into models poses difficulties in representation and parameterization. In addition, the lack of a unified, validated theory of human behavioral change in response to an infectious disease, and limited collaboration with behavioral and social scientists, have isolated epidemic models from adequate incorporation of behavioral mechanisms (13). Acquiring reliable and detailed data on behavior and validating behavioral mechanisms are also challenging (14). Consequently, many epidemic models are not coupled with behavioral models or are limited in the extent to which they integrate change in human behavior (10). We argue that new developments in artificial intelligence (AI), specifically generative AI, are changing the behavioral modeling landscape, with major implications for epidemic modeling.

The field of generative AI has witnessed remarkable growth in recent years, revolutionizing various domains of science and research (15-17). Generative AI refers to AI with the ability to generate new content, such as images, text, music, and videos, by learning patterns and structures from existing data (18). Generative AI models (e.g., GPT-3.5, DALL-E) are first trained on a diverse corpus of data (e.g., books, music, articles, images, and websites), thus increasing their usability and range of applications (19). We build on this promising idea and leverage generative AI's power to create more realistic representations of human behavior for epidemic modeling. We hypothesize that generative AI can help incorporate human behavior in epidemic models by offering a modeling paradigm that is radically different not only from conventional compartmental models, but also from rule-based individual-level models (often referred to as agent-based models, or ABM).

In this paper, we present a generative agent-based model (GABM) of an epidemic that incorporates changes in human behavior by utilizing generative agents as representative proxies. Unlike conventional ABMs, this new paradigm has empowered each agent by generative AI giving them the ability to reason and decide based on available information without modelers enforcing any decision-making rule—as Figure 1 illustrates. Our proposed paradigm introduces agents with distinct personas, representing the society, that capture various characteristics capable of influencing human responses to an evolving epidemic from an infectious virus. Similar to residents in a town, these agents can go through daily routines (such as reading the daily news) and make decisions (such as whether to go to work). Crucially, the agents' decisions regarding whether to go to work are influenced by their individual personalities (20), information about the virus, their own perceived health, and the perceived risks of infection derived from multiple information sources, including daily case counts reported in news outlets. Their decision-making process is powered by a pretrained large language model (LLM) such as ChatGPT. Leveraging an LLM, these agents possess realistic personas that, while varying across individuals, represent different groups of people in society.

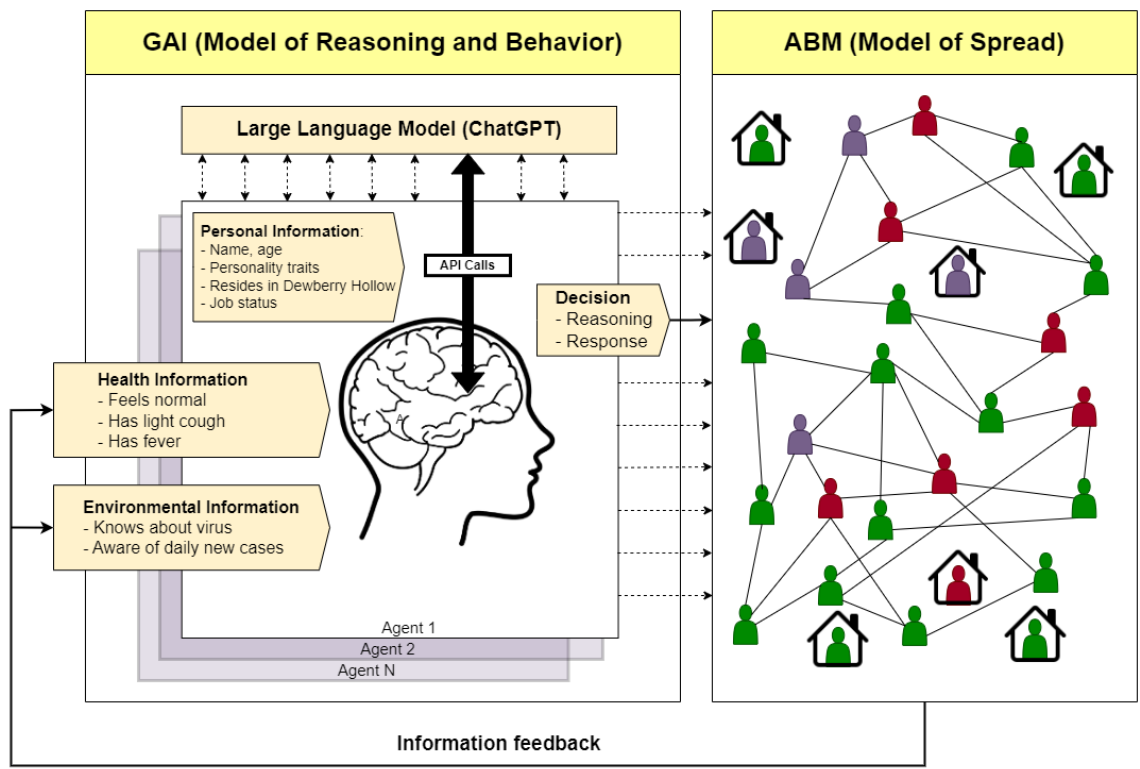

Fig. 1. Generative agent-based modeling provides reasoning and decision-making ability for each agent through closing a feedback between an AI-based large language model of reasoning (GAI) and a simulation model of spread (ABM). In the ABM section, green represents people susceptible to the virus, red infected people, and gray people who have recovered from the virus.

Through several simulation experiments, we present compelling evidence that configuring our model to closely mimic real-world conditions yields agents who are risk-responsive, self-isolate as the number of cases grows, and quarantine when they feel sick. Moreover, the agents consider age-related risks and their reasoning and decisions are affected by their personalities. Collectively, the agents' interactions result in emerging patterns akin to many real-world observations. Notably, the emerging patterns qualitatively mimic the multiple waves observed in recent pandemics followed by a period of endemicity. Our agents collectively flatten the epidemic curve and decrease the total number of cases.

This research represents a significant advance in the field of epidemiological modeling resulting from allowing AI to integrate epidemic modeling with the social and behavioral sciences. This integrated approach allows for the representation of human behavior and its interconnectedness with the spread of the disease, fostering a dynamic interaction between the two. This progress serves as a crucial step toward addressing the challenges associated with incorporating human behavior in epidemic models, with applications in many other complex social system modeling problems.

## II. Materials and Methods

Conceptual Model

As Fig. 1 shows, the model is a combination of two parts: GAI to model each agent's reasoning and behavior, and ABM to model the spread of the disease. At the beginning of each time step, agents begin in the GAI portion and decide whether to go outside, in which case their interactions with others for that day will be modeled by ABM.

In the GAI sector, powered by ChatGPT, each agent is fed a prompt that includes the agent's name, age, traits the agent possesses, basic biography, and relevant memories. The information comprising the relevant memories is set based on the experimental condition. For instance, it can include the agent's symptoms (if any) or information about the percentage of sick individuals in the town (see "Experimental Design" below for details). Then, at each time step, agents are asked whether they should stay at home for the entire day and their reason. For those who decide to leave their home, the ABM portion will have each agent individually interact with a number of unique agents equal to the model's contact rate, which can result in disease transmission between a susceptible person and an infected person. Once all agent interactions are carried out, the time step increments by one, and health statuses are updated.

Experimental Design

Our experimental design includes three main conditions: base run, self-health feedback, and full feedback. During the base run, agents are informed about the town, their personalities and age, and the fact that they go to work to earn a living. In this condition, there is no information feeding back to the agents, and even though they can decide whether to stay home and consequently not interact with others, we expect the model to replicate basic ABM outcomes. In the self-health feedback condition, in addition to the base run information, agents are informed about health symptoms they are experiencing (if any), which can potentially cause them to self-quarantine by staying at home. We hypothesize that some agents will practice self-quarantine based on information about their symptoms, which should in turn decrease the infection rate. In the full feedback condition, in addition to the self-health feedback, agents read daily news that includes information about the percentage of people in the town reported to have Catasat symptoms. We hypothesize that some agents will practice self-isolation, a behavior that correlates with information about the spread of the disease in the town, and as result patterns for the spread of the virus resemble oscillatory patterns. For all three types of model simulations, each agent provides a daily decision for each time step of whether to go to work, and their reasoning behind the decision. In addition, for sensitivity analysis, we repeat our n=1000 for different infectivity and contact rate values that corresponds to R0 of 2, 2.5, and 3. The results are reported in the Appendix.

Coding

We use a Python script to operationalize our GABM approach. The LLM used is gpt-3.5-turbo-0301 accessed via OpenAI API calls. Agents and the world they live in were defined using a Python library called Mesa (26). Their names were selected using the names-dataset 3.1.0 Python library (https://github.com/philipperemy/name-datase), and ages were randomly selected integers from 18 to 64. Agent traits are based on the Big Five traits typically used in psychology (20), and we gave a 50% chance of a positive versus negative version of each trait for each agent.

In the base run, the agent has the following relevant memory: "[agent's name] goes to work to earn money to support [agent's name]'s self." For the own health feedback case, agents' relevant memories include the base run memory and one of these three health strings: "[agent's name] feels normal," "[agent's name] has a light cough," and "[agent's name] has a fever and a cough."

The disease duration is assumed to be 6 days. The first two days of infection are assumed to be asymptomatic (agents feel normal) but infectious, while in days three and six, a light cough, and days four and five, a fever and cough are experienced. Upon recovery the agents feel normal and are immune.

For the societal health information feedback, the following string is given to the agent: "[agent's name] knows about the Catasat virus spreading across the country. It is an infectious disease that spreads from human to human contact via an airborne virus. The deadliness of the virus is unknown. Scientists are warning about a potential epidemic. [Agent's name] checks the newspaper and finds that [X]% of Dewberry Hollow's population caught new infections of the Catasat virus yesterday," where X is the percentage of the population with yesterday's new daily active cases. New daily active cases are counted by the number of agents who are in day 4 of being infected.

In the first round of simulation, we implement these three conditions with a town of n=100 individuals and an average contact rate of 5 contacts per individual. The virus infectivity (infection probability given a contact between an infected and a susceptible) is set at 0.1, which gives an average R0 of 3. We repeat each of the three conditions ten times (total of 30 runs), each run taking about two hours to complete. In the second round, we scale up the analysis for n=1000 individuals, and conduct two experiments per condition of R0 of 2 (infectivity is set to 0.0833 and contact rate of 4, R0 of 2.5 (infectivity is set to 0.0833), and R0 of 3 (total of 7 runs). Each run takes about 80 hours to complete.

## III. Results

Given the computational intensiveness of this study, we first conduct our simulation experiments with a population of 100 agents. We use the pseudonyms of Catasat and Dewberry Hollow for the virus and the town's names, respectively, to avoid any unknown biases in LLM. When information about the virus is provided, it is specified that Catasat is an airborne human-to-human infectious virus with unknown deadliness and that scientists are warning about a potential epidemic.

Figure 2 shows the results for this population under three experimental setups, each one conducted 10 times (details in the Materials and Methods section). In the first condition (Fig. 2A), there is no information feedback to the agents, and we expect the model to replicate the common observation in conventional susceptible, infectious, and recovered (SIR) models (compartmental or ABM) of a single bell-shaped curve of daily cases and constant mobility. Fig. 2A shows average daily cases from 10 simulation experiments (80% CI), average mobility (80% CI), and 4 samples of individual runs, all depicting the expected outcomes. The epidemic continues until almost everyone is infected. Fig 2B reports results from the second experimental condition, in which agents are informed about their own health and, based on this new information, they can alter their decision regarding whether to go outside. The results show a decline in mobility due to self-quarantine, and the subsequent decline in the maximum peak of the epidemic.

Fig. 2C shows the full feedback condition, in which agents learn about the town's cases in addition to discovering their own health-related feelings. As depicted, agents react to the growing number of cases, self-isolate, and as a result, the number of cases declines. Generative agents provided multiple reasons for choosing to stay home, including perceived higher risks and the presence of symptoms. For example, the reason Carol (an agent) chose to stay home at a particular time is indicated as: "Carol has a light cough and there is a potential epidemic of an unknown deadly virus spreading through human contact. Additionally, 0.7% of Dewberry Hollow's population caught new infections of the virus yesterday. Going to work could increase the risk of exposure to the virus and potentially spread it to others."

Generative agents who chose to leave home for work provided reasons such as the need to earn money. For example, the same agent Carol gives the following reason for going to work at a particular time: "The percentage of new infections in Dewberry Hollow is relatively low and Carol needs to go to work to earn money." (Fig. 4 gives more examples of reasons.) This case demonstrates that without modelers imposing any decision-making rules on how to behave based on the available information, agents' reasoning (which is based on an LLM) influences their decisions. Collectively, agents flatten the curve of the epidemic as they react. The four individual sample runs on the right side also show the emergence of epidemic waves, which can range from a single wave to multiple waves of different sizes.

Fig. 2D compares several measures across the experiments. It shows that generative agents with information about prevalence are able to bring down the number of cumulative cases and the maximum number of cases by lowering their own mobility, while their reactions double the epidemic duration.

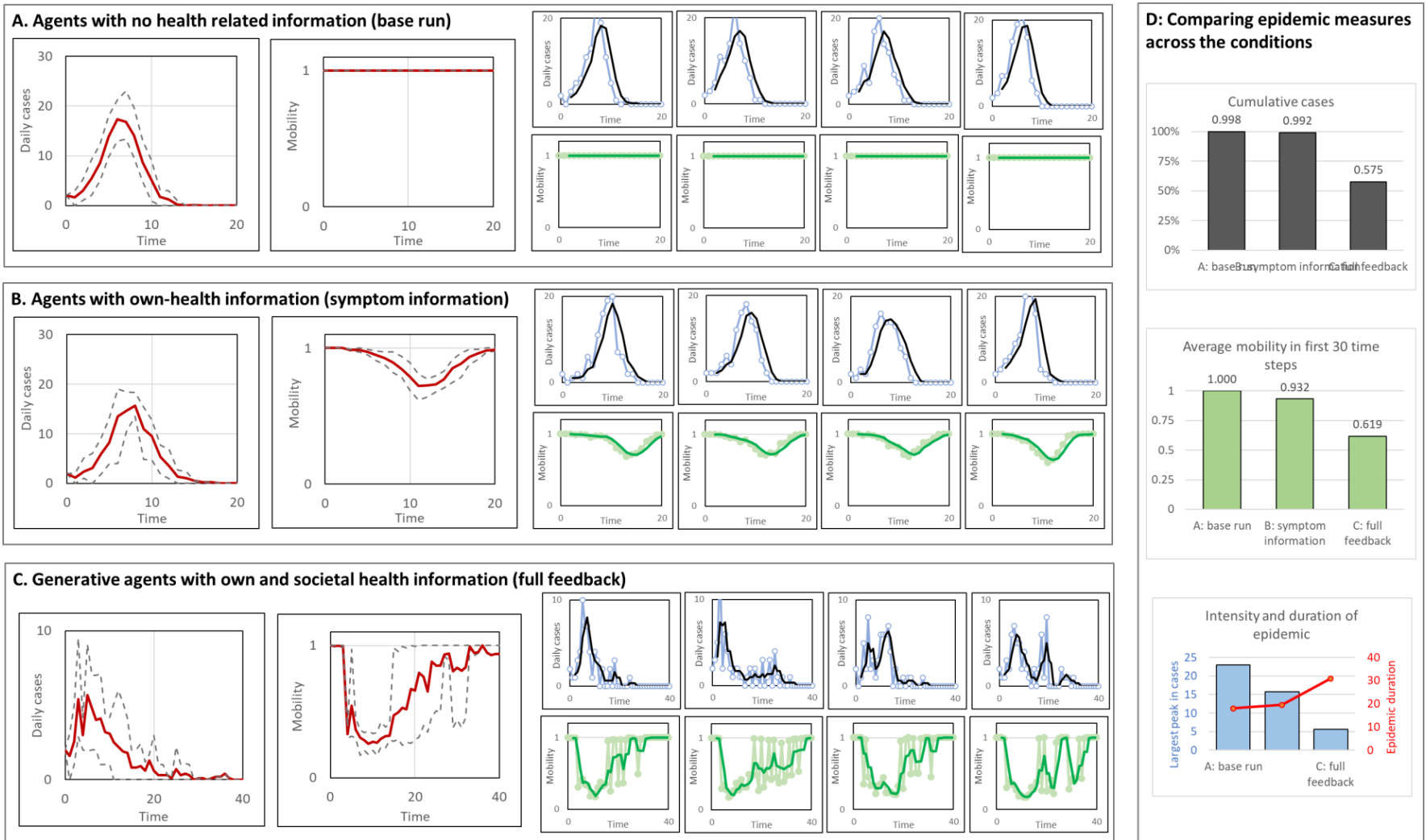

Fig. 2. Comparison of simulation results from three experimental conditions with a 100-agent population. Panels (A-C): average daily cases and mobility (and their 80% confidence interval with dashed line) as well as four samples of simulation (darker lines are 3-day moving average) for (A) the first condition with no health information to agents, (B) when own-health information is provided, and (C) when own health and societal health information is provided. Panel (D) compares cumulative cases, average mobility, largest peak in cases, and epidemic duration across the three experimental conditions.

We repeat the full feedback experiment for a population of 1,000 agents for three different conditions of infectivity and contact rate to represent different initial reproductive numbers ($R_0$) of 3, 2.5, and 2. Fig. 3 shows the results. Panel A shows daily cases, depicting different modes of outcome from one wave ($R_0=3$) to multiple waves ($R_0$ of 2 and 2.5). Panel B presents cumulative cases, showing that herd immunity is achieved in a much lower number of total cases than the entire population. Panel C shows mobility over time, confirming that we have a more restrictive response in higher $R_0$s. Finally, panel D depicts the correlation between past daily cases that were communicated with the agents and the collective decision of going out, presented by the percentage that decide to go to work. The observed relation follows a negative exponential relationship and is qualitatively consistent with several empirical studies of how humans responded to change in disease prevalence or daily death rate (9, 21). To examine the robustness of the results, we ran the model seven more times for the population of 1,000 individuals, with different $R_0$ values (see Supplementary Material). The results are qualitatively consistent.

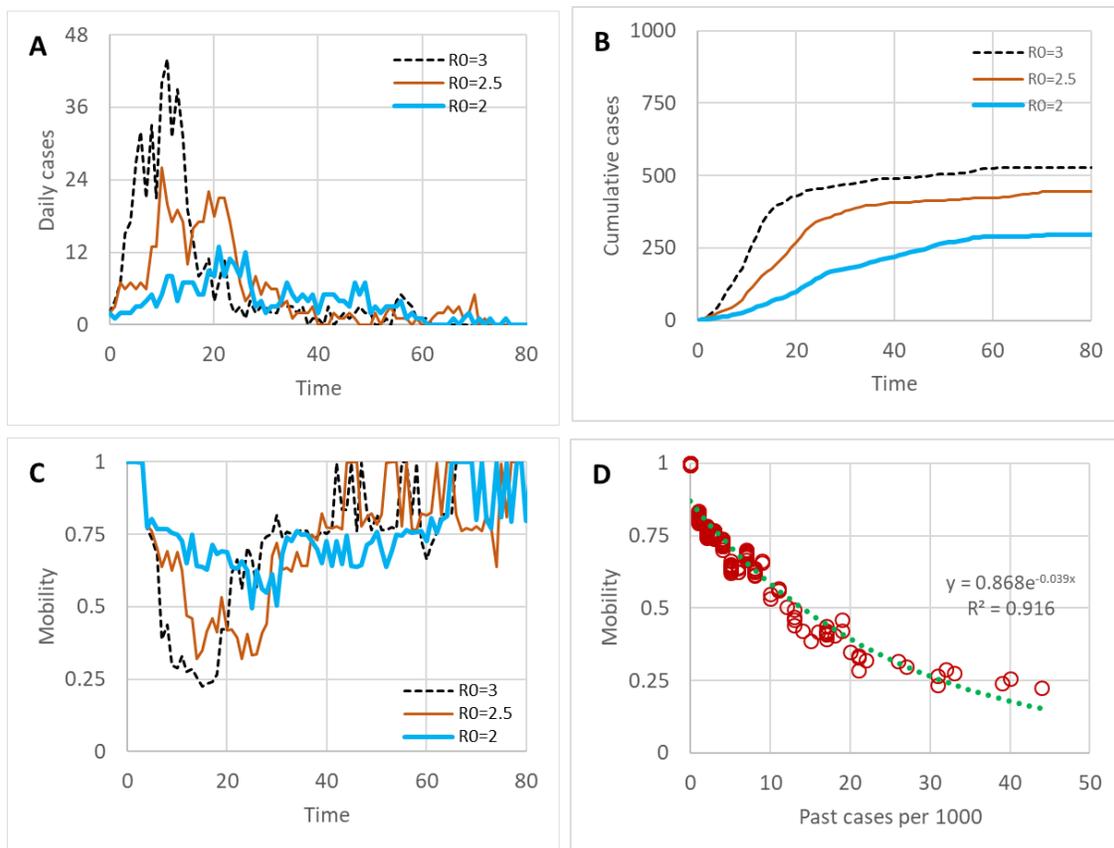

Figure 3. Results from three simulation experiments of different initial reproductive numbers with 1,000 generative agents who receive feedback about daily cases and their own health symptoms. A) daily cases, B) cumulative cases, C) mobility, and D) agents' responsiveness to risks depicted by association between past case and current mobility.

Finally, we conduct an individual-level analysis of the agents' decisions to stay home. Fig. 4 and Table 1 report the results. Fig. 4A shows a prompt sample used to prime one of the agents. Panel B demonstrates the distribution of days during which agents stayed home. In a 68-day simulation, of the simulation, about 1% of agents did not stay home on any days, while more than 8% remained at home for five days. Panel C depicts three agents' personalities, their decisions, and a sample of their reasons for staying home or going outside, indicating agents provided diverse reasons for their decision.

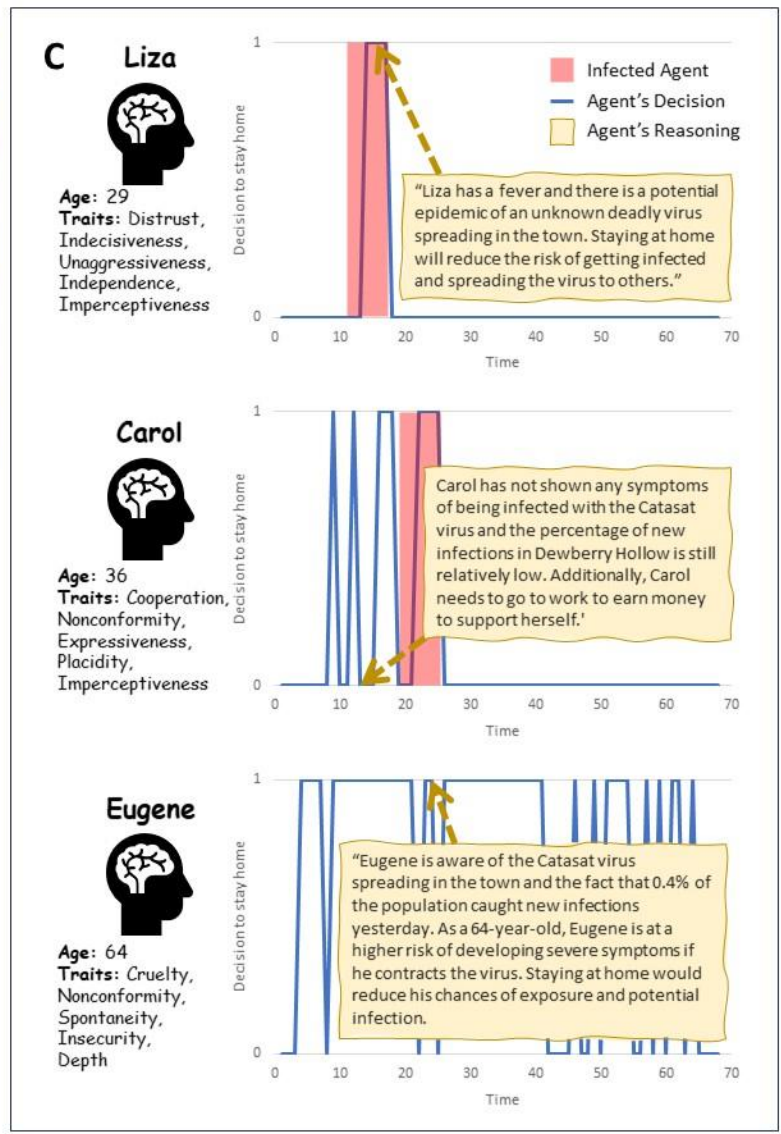
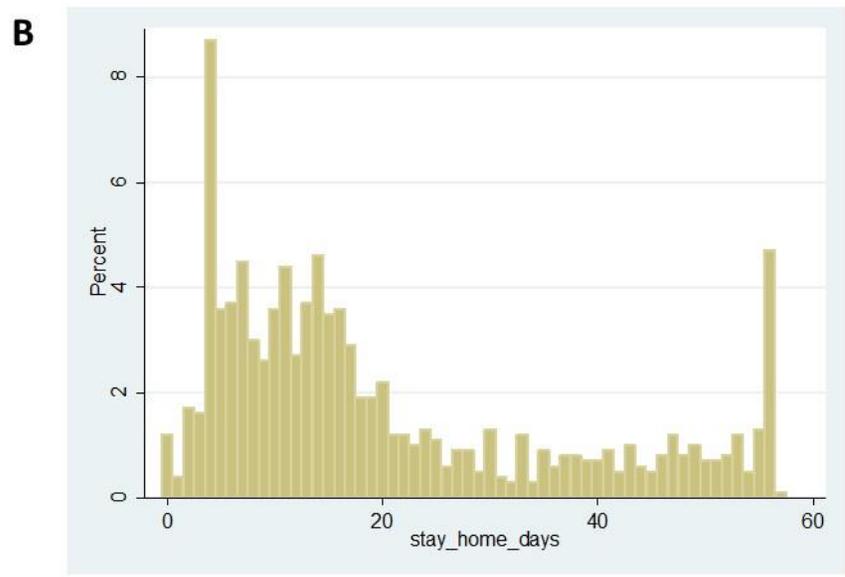

Figure 4. Agents' responses to their prompts. A: a prompt sample; B: the distribution of responses as stay-at-home-days (out of 68 days); C: a sample of individuals' personalities, reactions, and their reasoning.

In order to examine systematically whether health information about agents and their town influences their behaviors, we conduct several regression analyses (reported in Table 1). The dependent variable is a binary variable of an agent's decision to stay home (equal to 1 if the agent stays home and 0 otherwise). The first three regressions are fixed effect logit models that investigate the impact of agents' own health and societal health information on their decisions. As regression 1 shows, those with a light cough and both a fever and cough are more likely to stay home than those who feel "normal." The improvement in both pseudo R-squared and BIC from regression 1 and 2 shows the importance of societal health information for predicting agents' decisions. In regression 3, adding the squared of societal health information enhances both BIC and pseudo R-squared, indicating that societal health information has a nonlinear relationship with agents' decisions. In other words, agents respond to societal health information, but their responses diminish over time. Regression 4, a random effect logit model, examines the impact of different personality traits, age, and gender on agents' response. The direction of the main independent variables is consistent with regressions 1–3.

Table 1 – Statistical analysis of agents' decision to stay home (binary dependent variable)

| Regressions | Regression 1 (N=66,960) | Regression 2 (N=66,960) | Regression 3 (N=66,960) | Regression 4 (N=67,776) |
|---|---|---|---|---|
| Own Health | | | | |
| Light Cough | 5.69*** (0.14) | 5.75*** (0.16) | 5.40*** (0.16) | 5.60*** (0.16) |
| Fever & Cough | 5.42*** (0.13) | 5.26*** (0.15) | 4.94*** (0.15) | 5.13*** (0.15) |
| Societal Health | | 1.59*** (0.02) | 3.97*** (0.05) | 3.95*** (0.05) |
| Societal Health$^2$ | | | -0.65*** (0.01) | -0.65*** (0.01) |
| Agreeableness | | | | -0.11 (0.17) |
| Conscientiousness | | | | -0.70*** (0.17) |
| Surgency | | | | -0.26 (0.17) |
| Emotional stability | | | | -0.59** (0.17) |
| Intellect | | | | -0.87*** (0.17) |
| Age | | | | 0.01 (0.01) |
| Gender | | | | -0.84*** (0.17) |
| Fixed effect | × | × | × | |
| Random effect | | | | × |
| Pseudo R$^2$ | 0.12 | 0.41 | 0.46 | |
| BIC | 52,359 | 35,227 | 31,949 | 39,793 |

Note: ***: p<0.001, **: p<0.01

# IV. Discussions

In this paper, we offer a new epidemic modeling approach to incorporate human behavior in infectious disease models by leveraging generative AI, specifically LLMs. In this paradigm of individual-level epidemic modeling, each agent is empowered by AI to make decisions (here, whether to go outside) by correlating different pieces of contextual information (such as their personality traits, health status, and the prevalence of the disease) with the common knowledge that exists in an LLM (here, ChatGPT). In this approach, rather than a modeler formulating decision-making rules and estimating their corresponding parameter values for each agent, she empowers each agent with a reasoning power by connecting them to a well-performing LLM. We show that without the modeler imposing decision choices (i.e., decisions as exogenous inputs), or how to make decisions (i.e., decision rules), agents can make decisions that are consistent with how humans behave in the real world. For example, in our case, they are more likely to decide to self-isolate when cases rise, quarantine when they feel sick, or stay home if they have a more risk-averse personality or are elderly. Through interactions of these generative agents, two emergent behaviors are observed: the agents are collectively able to flatten the curve of the epidemic; and the system recreate various modes of an epidemic, including multiple waves and continuing endemic states.

We demonstrate the results in three steps by performing various simulation experiments coupled with a conventional agent-based model with an LLM through API calls. In the base-run scenario, we show that in the absence of feedback about the disease, self-health, and societal health, the generative agents behave similar to rule-based agents in classic ABM frameworks and SIR compartmental models. In the next step, informing agents about their own health at the beginning of each time step, we observe that agents with symptoms are more likely to decrease their mobility. Most agents with symptoms of a fever and cough quarantine themselves by staying home. As a result, agents are able to slow the spread of the disease. Finally, when agents are primed with societal health information, news about the epidemic, and the daily active case count in their simulated town, they are able to flatten the curve of the epidemic substantially in their town by self-isolating. Individual-level analysis shows variation in the decisions of agents and different responsiveness to news about the virus, resembling a real-world situation.

This research contributes to the literature on infectious disease modeling (22) by proving a new epidemic modeling approach. While several previous studies stressed the importance of coupling disease models (such as compartmental models or ABMs), few have formally pursued this path (23, 24). Our research contributes to this body of literature by providing a novel way for closing the feedback loop between disease and human behavior. This approach, while resonating with dynamic approaches that endogenously formulate human behavior (25), differs from conventional dynamic modeling approaches by relying on LLMs to represent human responses to the state of the system rather than on explicit mathematical formulations and parameterizations of human behavior.

Beyond creating a new method of epidemic modeling, this study contributes to the literature on complexity and complex system modeling by providing a new approach to incorporating human behavior in simulation models of social systems. Identifying, formulating, and parametrizing human responses in complex systems are always challenging; in the generative agent approach, modelers can rely on LLMs to represent human response to change in the state of the system.

This has the potential to change how complex systems are modeled with their human elements, leading to powerful dynamic models that more accurately represent human responses at the individual level. This accuracy comes from the fact that pretrained AI programs are fostered by large volumes of textual data about how humans behave under similar conditions.

In addition, this work contributes to the evolving field of GAI. Over the past year, generative artificial intelligence has begun to be disseminated into all parts of society. In terms of behavioral modeling, Park and colleagues (17) have provided a groundbreaking contribution by developing and using a memory architecture for their models. In our model, due to computational complexity, a higher number of agents, and a longer period of simulation, we had to avoid the computational intensity that comes with the use of memory. Thus, our work differs architecturally by using prompt engineering, which is equivalent to a daily reminder for each individual. Nevertheless, technological advancements and higher computational powers can facilitate the use of memory architecture in generative agents and improve GABMs.

A major limitation of this study is related to the fact that the field of generative AI is in its infancy and still developing. Our models are relatively resource-intensive, costly and relatively time-consuming to run. As of June 2023, an epidemic model of 1,000 agents can run for more than 90 hours with 32 GB of RAM CPU of program runtime costing about $20 per run, primarily due to the tens of thousands of application program interface (API) calls made to OpenAI's servers, in addition to other hardware costs. However, as LLMs improve in cost and speed in the coming years, there is hope that GABM's computational expense and time can be reduced. There is already some hope with the advent of local-running LLMs. We invite modelers from different disciplines— from economics, political science, sociology, and ecology to epidemiology and health policy—to explore this avenue of modeling and contribute to more realistic representations of human reasoning and behavior in complex systems.

## V. Acknowledgments

**Funding:** This research is funded by US National Science Foundation, Division of Mathematical Sciences & Division of Social and Economic Sciences, Award 2229819.

**Author contributions:** Conceptualization: RW, NG; Methodology: RW, NH, NG; Coding: RW, AM; Data gathering and data curation: RW; Investigation: RW, NH, AM, NG; Funding acquisition: NH, NG; Supervision: NH, NG; Writing: RW, NH, AM, NG;

**Competing interests:** Authors declare that they have no competing interests.

**Data and materials availability:** All data, data-processing and analysis code, as well as the full model, its associated files, and results files, are available online at https://github.com/bear96/GABM-Epidemic. Full model description and supplementary analysis are available in the Supplementary Materials.

# Supplementary Materials

Operational Details
In our proposed model, we use the Mesa Python library for agent-based modeling (ABM) to create two components: the Citizen and the World, both of which are Python classes. Below are more details about these major components.

Citizen Architecture
The Citizen class plays a significant role in the simulation. It encompasses a range of methods that facilitate the agent's interaction with the simulated environment, enable decision-making based on its current state, and simulate the infection dynamics.
Table S1 describes the high-level algorithm of one agent going through their day. To implement the algorithm, we design the Citizen class to have attributes and functions that create the make-up of agents. Table S2 displays each attribute of the Citizen class; the table mentions attribute name, attribute description, and whether the value of the attribute state is changing dynamically or remaining statically through the simulation. Having explored the attributes of the Citizen class in Table S2, Table S3 contains the Citizen class methods that play a crucial role in the overall functionality of the program by utilizing the attributes. Table S3 provides the method name, a brief description of its purpose, and details on which other functions it is utilized in.

World Architecture
The world class simulates a world with agents and their interactions, considering contact rates, agent locations, infections, recoveries, and checkpoints for model state persistence. Table S4 outlines the sequential progression of a day in the world, providing a detailed breakdown of each step along the way. The world architecture has attributes and functions that help implement the algorithm in Table S4. Table S5 displays each attribute of the world class while Table S6 indicates the class methods used in conjunction with the attributes.

Prompt Engineering
     The prompt fed into turbo gpt-3.5-turbo-0301 is a system message. System messages are used to tune the behavior of ChatGPT's responses. System messages, compared to user messages, enhance ChatGPT's ability to maintain character and follow rules over a prolonged conversation. Hence, when asking ChatGPT to roleplay, it was a natural fit. Secondly, to parse ChatGPT's responses, it is important that it responds in a structured manner. We used system messages for three purposes: imbuing a persona, providing relevant context and output formatting.
     Fig. S2 shows the prompt we fed into ChatGPT. Imbuing the agent's persona occurs by telling ChatGPT the agent's name, the agent's age, and the agent's traits. Additionally, a basic agent bio is given: the agent happily lives in the town of Dewberry Hollow and has a job requiring the agent to go to the office for work everyday. This portion of the bio is to incentivize the agent to naturally want to leave home. The second portion of the prompt is to provide relevant information to the agent. The initial piece of relevant information, when implemented, is the health feedback information. The next part of relevant information, when enabled, the agent is made aware of the Catsat virus and the percentage of new active cases found in Dewberry Hollow. Lastly, within the relevant information section, the agent is told that it goes to work to earn money to support oneself. The third portion of the prompt is the question of whether the agent should stay at home for the entire day. The entire day portion of the question emphasizes

the point that the agents that remain at home will be in isolation. The last part of the prompt is to help ChatGPT to output a reasoning and response in a format that we can dissect.

Out of 68,000 data points fed into the prompt, we found that less than 0.33% of responses disobeyed the prompt orders of providing a response of either yes or no. Those non-conforming responses are defaulted to a "no" value.

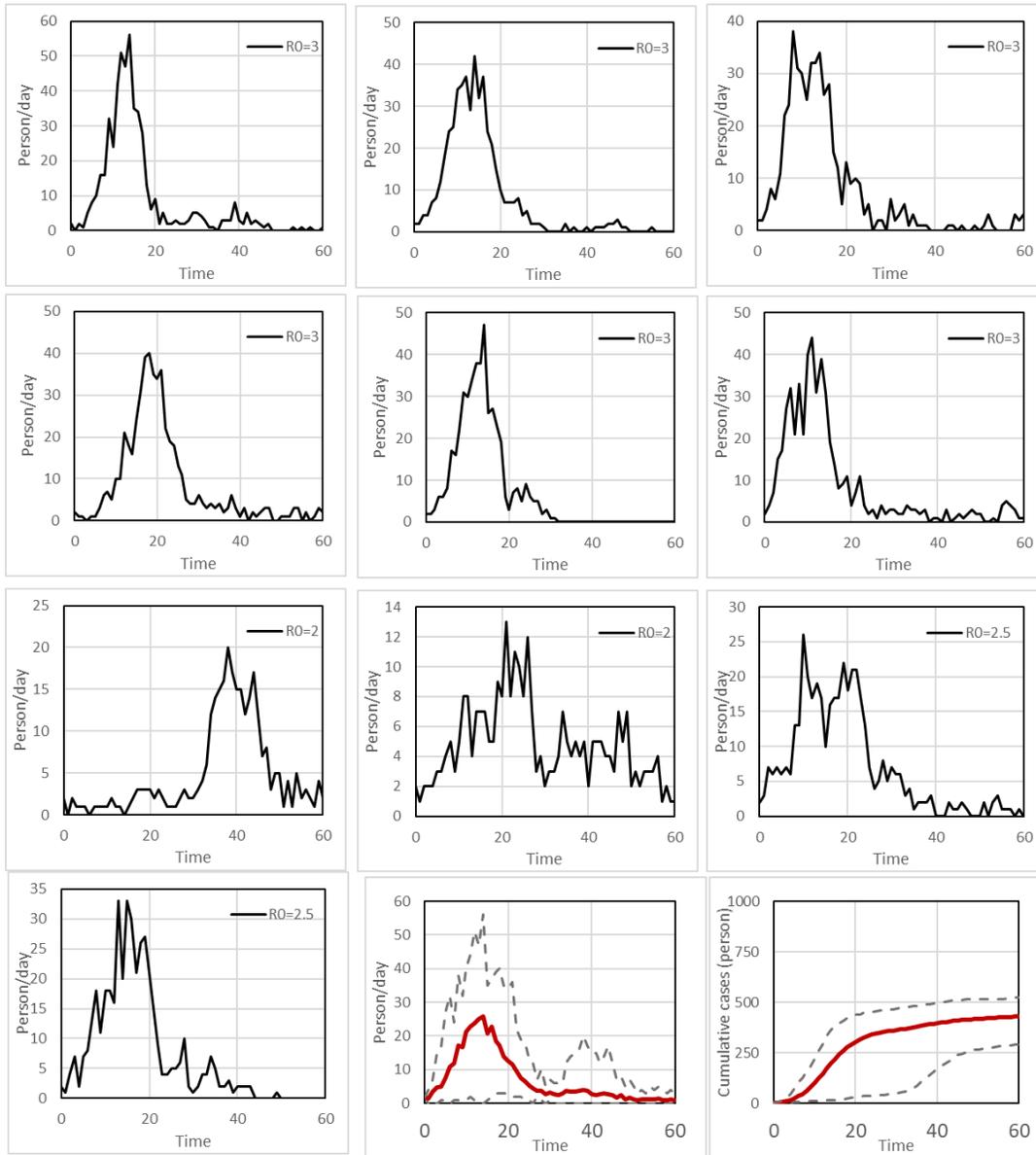

**Fig. S1.**
Results from 10 simulation experiments with the full model and different R0s. The final two panels show the average results with dashed lines showing minimum and maximum of observed cases.

> You are [agent's name]. You are [agent's age] years old.
> Your traits are given below:
> [agent's traits]
> Your basic bio is below:
> [agent's name] lives in the town of Dewberry Hollow. [agent's name] likes the town and has friends who also live there. [agent's name] has a job and goes to the office for work everyday.
> I will provide [agent's name]'s relevant memories here:
> [agent's health feedback]
> [agent's name] knows about the Catasat virus spreading across the country. It is an infectious disease that spreads from human to human contact via an airborne virus. The deadliness of the virus is unknown. Scientists are warning about a potential epidemic.
> [agent's name] checks the newspaper and finds that
> [X]% of Dewberry Hollow's population caught new infections of the Catasat virus yesterday.
> [agent's name] goes to work to earn money to support [agent's name]'s self.
> Based on the provided memories, should [agent's name] stay at home for the entire day? Please provide your reasoning.
> If the answer is "Yes," please state your reasoning as "Reasoning: [explanation]."
> If the answer is "No," please state your reasoning as "Reasoning: [explanation]."
> The format should be as follow:
> Reasoning:
> Response:
> Example response format:
> Reasoning: [agent's name] is tired.
> Response: Yes
> It is important to provide Response in a single word.

**Fig. S2.**
The daily ChatGPT prompt asked to each agent.

**Table S1.**
Daily algorithm for an agent

| Step Number | Action | Decision Variables, Inputs, and Outcomes |
|---|---|---|
| 1 | Determine if the agent stays home or goes out | Inputs: provided context, potentially including different types of feedback information. Decision model: entrusted to ChatGPT to analyze the input context and determine the agent's choice. Outcome: *stay home* or *go out* |
| 2 | Update agent's location | Inputs: the decision made in step 1. Update: if *stay home*, set location to home. If *go out*, set location to *grid* |
| 3 | Enable interactions with other agents | Condition: if the agent chose *go out* in step 1, it interacts with other agents. Interactions: agent interacts with up to X amount of other agents, where X is the *contact_rate*. Outcome: list of agents to interact with |
| 4 | Assess possibility of infection transmission | Inputs: health condition of interacting agents in the list from step 3. Condition: if an agent is *infected* and contacts another agent who is *susceptible* and infection rate>randomly generated number ∈ [0,1]. Outcome: update state to *to_be_infected* if the condition is met |
| 5 | Update health condition attributes | Inputs: health condition from step 4. Updates: if health condition is *to_be_infected*, update to *infected*, set number of days infected to 0 |
| 6 | Monitor and manage infection progression | Inputs: health condition and number of days infected. Conditions and Updates: if *infected*, increment number of days infected by 1. If *infected* and the number of days exceeds the healing period, update to *recovered* and reset the number of days infected to *None* |

**Table S2.**
Citizen class attributes

| Attribute name | Attribute description | Attribute state |
|---|---|---|
| unique_id | Used for scheduler step function and data collection | Static |
| name | The name of the citizen | Static |
| age | The age of the citizen (age ∈ [18, 65) and age ∈ Z) | Static |
| traits | The traits of a citizen selected randomly from each of the Big Five traits (X) | Static |
| location | Determines whether the citizen is staying at home or going out on the grid. Going to grid and going to work are synonymous in the code base because we leverage Mesa's Library to create a virtual grid for agents who go outside. | Dynamic |
| pos | The Mesa library requires agents to have a pos (position). This is not used in our code beyond setting a position for each agent on the grid. This could be useful for animation of the simulation in the future. For now, it can be functionally be ignored in the code | Static |
| Health_Condition | Health condition of the agent can be "Susceptible", "To_Be_Infected", "Infected", or "Recovered" | Dynamic |
| day_infected | Stores the number of days the citizen has been infected | Dynamic |
| agent_interaction | Stores citizen objects with which the current citizen will interact with in the current timestep | Dynamic |
| width, height | Defines the width and height of the simulated world. | Static |

**Table S3.**
Citizen class methods

| Method name | Method description | Where Method is used |
|---|---|---|
| __init__() | Initializes the Citizen object by setting its attributes | __init__() of World Class |
| get_health_string() | Returns a descriptive string representing the citizen's health condition based on the number of days infected. It provides different health descriptions for different infection stages | get_response_and_reasoning() |
| ask_agent_stay_at_home() | Checks if response provided by LLM is a "yes" or "no" and returns True and False respectively | decide_location() |
| get_response_and_reasoning() | Provides a prompt to LLM and parses the output. Returns response and reasoning by the LLM | ask_agent_stay_at_home() |
| decide_location() | If ask_agent_stay_at_home() returns True, agent's location is set to "home" else it is set to "grid" | prepare_step() |
| add_agent_interaction() | Creates a list of agents for interaction in the world | decide_agent_interactions() of World class |
| interact() | Makes agents interact with other agents in the agent's agent_interaction list | step() |
| infect() | Infects another agent with respect to a probability threshold and health status | interact() |
| prepare_step() | Prepares the agent for the eventual step() function by deciding on its location first | step() of World class |
| step() | Agent interacts with all the agents in their interaction list. | step() of World class |

**Table S4.**
Algorithm for a day in the World

| Step Number | Description |
|---|---|
| 1 | Initialize the simulation by creating the world and its agents |
| 2 | At the beginning of each time step, agents independently decide whether to stay home or go outside |
| 3 | For agents who decide to go outside, calculate potential interaction partners by identifying agents also present on grid |
| 4 | Allow agents to interact with their potential partners, potentially infecting them based on the *infection_rate* and *health_condition* |
| 5 | Update the state of the agents and the world based on their actions and interactions during the day |
| 6 | Calculate the *day_infected_is_4* by tallying the number of agents who have reached the fourth day of their infection |
| 7 | Save the entire world, including agents and their attributes, into a local file as a checkpoint at the end of each time step |
| 8 | Check for an early stopping condition: if there are no infected agents for two consecutive days, save the final checkpoint and halt the simulation |
| 9 | Provide the *day_infected_is_4* as feedback to agents at the beginning of the next time step |

**Table S5.**
World class attributes

| Attribute Name | Attribute Description | Attribute State |
| --- | --- | --- |
| initial_healthy | The initial number of healthy (susceptible) agents | Static |
| initial_infected | The initial number of infected agents | Static |
| contact_rate | The maximum number of interactions an agent can have per step | Static |
| step_count | The number of time steps to simulate | Static |
| offset | Helper variable to enable loading checkpoints | Static |
| name | Name for saving checkpoints and simulation outputs | Static |
| height, width | Dimensions of the world grid | Static |
| grid | The grid where agents exist. This is functionally not used | Static |
| current_date | The current date in the simulation. This has no functionality, but is used simply for logging purposes | Dynamic |
| total_contact_rates | Stores the total number of contacts in each time step | Dynamic |
| track_contact_rate | List that tracks the total contact rate over time steps | Dynamic |
| day_infected_is_4 | List to track the number of agents infected for 4 days | Dynamic |
| list_new_cases | List to store the number of new cases for each day | Dynamic |
| daily_new_cases | The number of new cases for the current day. It is initially set equal to the number of agents initially infected (initial_infected) | Dynamic |
| infected | The total number of infected agents | Dynamic |
| agents_on_grid | List of agents currently on the grid | Dynamic |
| max_potential_interactions | Maximum number of potential interactions an agent can have with other agents who are also on grid | Dynamic |
| schedule | Scheduler for agent activation. This is provided by Mesa (mesa.time.RandomActivation) | Static |
| datacollector | Collects data during the simulation. This is provided by Mesa (mesa.DataCollector) | Dynamic |

**Table S6.**
World class methods

| Method Name | Method description | Where Method is used |
|---|---|---|
| __init__() | Initializes the World object and sets its attributes. It also creates the grid, initializes agents, and sets up the data collector. | Called when an instance is created of the World class |
| distribute_agents() | Distributes agents on the grid randomly. | __init__() function of the World class |
| decide_agent_interactions() | Determines the interaction partners for each agent based on the contact rate. The goal is to ensure that interactions between agents are a two-way process | step() function of World class |
| step() | Includes important steps such as agent preparations, determining interaction partners, calculating the total contact rate, executing individual agent steps, and updating agent attributes and global infection statistics | run_model() of the World class |
| run_model() | Runs the model for the specified number of time steps. It collects data, performs model steps, saves checkpoints, and prints relevant information | Called in the script that runs the simulation |
| save_checkpoint() | Saves a checkpoint of the model to the specified file path | run_model() of the World class |
| load_checkpoint() | Loads a checkpoint from the specified file path | Called in the script that runs the simulation |

2320. Goldberg LR. An alternative "description of personality": The Big-Five factor structure. Journal of Personality and Social Psychology. 1990;59(6):1216-29.
21. Ghaffarzadegan N. Simulation-based what-if analysis for controlling the spread of Covid-19 in universities. PLOS ONE. 2021;16(2):e0246323.
22. Kermack WO, McKendrick AG, Walker GT. A contribution to the mathematical theory of epidemics. Proceedings of the Royal Society of London Series A, Containing Papers of a Mathematical and Physical Character. 1927;115(772):700-21.
23. Qiu Z, Espinoza B, Vasconcelos VV, Chen C, Constantino SM, Crabtree SA, et al. Understanding the coevolution of mask wearing and epidemics: A network perspective. Proceedings of the National Academy of Sciences. 2022;119(26):e2123355119.
24. Rahmandad H, Sterman J. Quantifying the COVID-19 endgame: Is a new normal within reach? System Dynamics Review. 2022;38(4):329-53.
25. Richardson GP. Reflections on the foundations of system dynamics. System Dynamics Review. 2011;27(3):219-43.
26. Kazil J, Masad D, Crooks A, editors. Utilizing Python for Agent-Based Modeling: The Mesa Framework2020; Cham: Springer International Publishing.
23